# Performance Evaluation of Trajectory Tracking Controllers for a Quadruped Robot Leg

Hossein Shojaei
*Dept. of Mechanical Engineering*
*K.N.Toosi University of Technology*
Tehran, Iran
h.shojaei1@email.kntu.ac.ir

Hamid Rahmanei
*Dept. of Mechanical Engineering*
*K.N.Toosi University of Technology*
Tehran, Iran
hrahmanei@mail.kntu.ac.ir

Seyed Hossein Sadati
*Dept. of Mechanical Engineering*
*K.N.Toosi University of Technology*
Tehran, Iran
sadati@kntu.ac.ir

*Abstract*—The complexities in the dynamic model of the legged robots make it necessary to utilize model-free controllers in the task of trajectory tracking. In This paper, an adaptive transpose Jacobian approach is proposed to deal with the dynamic model complexity, which utilizes an adaptive PI-algorithm to adjust the control gains. The performance of the proposed control algorithm is compared with the conventional transpose Jacobian and sliding mode control algorithms and evaluated by the root mean square of the errors and control input energy criteria. In order to appraise the effectiveness of the proposed control system, simulations are carried out in MATLAB/Simulink software for a quadruped robot leg for semi-elliptical path tracking. The obtained results show that the proposed adaptive transpose Jacobian reduces the overshoot and root mean square of the errors and at the same time, decreases the control input energy. Moreover, transpose Jacobin and adaptive transpose Jacobian are more robust to changes in initial conditions compared to the conventional sliding mode control. Furthermore, sliding mode control performs well up to 20% uncertainties in the parameters due to its model-based nature, whereas the transpose Jacobin and the proposed adaptive transpose Jacobian algorithms show promising results even in higher mass uncertainties.

## I. Introduction

Legged robots have caught the attention of researchers because of their exceptional capabilities to navigate through natural environments that are challenging for tracked and wheeled robots. Nonetheless, creating these robots leads to some disadvantages such as sustaining stability on uneven terrains, handling the numerous degrees of freedom, and coping with the under-actuated nature of the robots. Although some legged robots have been constructed, their performance is poor and is not comparable with their biological counterparts.

While the model-based controllers were originally designed for fixed-based manipulators, they have been extended to floating-based legged robots such as humanoid and quadruped robots in [1] and [2], respectively. Sliding mode control (SMC) is a popular model-based control algorithm used in quadruped robot locomotion control [3]. This control algorithm relies on a mathematical model of the robot dynamics and uses a sliding surface to ensure smooth and stable motion. The advantages of SMC include its ability to handle the nonlinear systems and disturbances, as well as its robustness to parameter variations. However, this control algorithm can be challenging to implement and requires accurate knowledge of the robot dynamics.

Model-free control algorithms of robots for locomotion, such as the transpose Jacobian control algorithm, have shown practical advantages by not relying on possibly inaccurate dynamic models, which have been applied to quadruped [4] and biped [5] robots. Model-free controllers typically require tuning of the feedback gains to ensure the stability of the system which could produce unstable behavior in the unstable or uncertain environments.

The contribution of this paper is to apply an adaptive transpose Jacobian control for a MIMO system, i.e. the quadruped robot leg, in the task of trajectory tracking. The PI-algorithm is used to adjust the proportional and derivative matrix gains of the transpose Jacobian. Also, the proposed control strategy is compared with conventional transpose Jacobian and sliding mode control while tracking a desired trajectory. The root mean square of the errors and control input energy criteria was used to measure the effectiveness of the proposed algorithm, compared with normal transpose Jacobian and sliding mode control. In addition, the proposed control method shows promising results compared to the other two methods in terms of parameter uncertainty and deviation from initial conditions.

The rest of this paper is organized as follows. In section II, the kinematic and dynamic models of the quadruped robot leg are presented. The model-based SMC and model-free transpose Jacobian control algorithms are introduced in section III, and then the adaptive transpose Jacobian algorithm is proposed. In section IV, the control input energy and root mean square of errors (RMSE) of the path tracking are presented as criteria for evaluating the performance of the controllers. Thereafter, the designed controllers are simulated in MATLAB/Simulink to follow the semi-elliptical reference path, in section V. Finally, the paper concludes with some hints and remarks as future works.

## II. Kinematic and Dynamic Modeling

In order to study the kinematics of the quadruped robot leg, the systematic approach of Denavit-Hartenberg is utilized to connect the frames to the leg links, to determine the position of the robot's foot tip. In Fig. 1, a simple model of the robot mechanism is presented, which expresses the position of the foot tip relative to the base framework {0}. In Fig. 1, the parameters $l_1$, $l_2$ and $l_3$ express the length of leg links.

The length of the links, as well as the parameters of mass and moment of inertia, are presented in TABLE I.

The Denavit-Hartenberg (DH) parameters, which describe the position of the foot tip relative to the base framework, are presented in TABLE II.

By employing the DH parameters, the resultant kinematic model of the robot leg is expressed as follows:

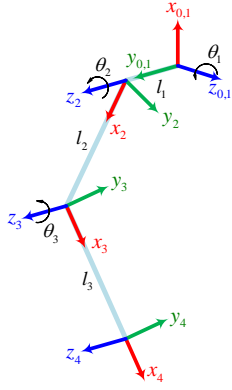

Fig. 1. Schematic of the robot's leg and coordinate frameworks.

$$\begin{cases} x = -l_1 S_{\theta_1} + l_2 C_{\theta_1} C_{\theta_2} + l_3 C_{\theta_1} C_{\theta_2+\theta_3} \\ y = l_1 C_{\theta_1} + l_2 S_{\theta_1} C_{\theta_2} + l_3 S_{\theta_1} C_{\theta_2+\theta_3} \\ z = l_2 S_{\theta_2} + l_3 S_{\theta_2+\theta_3} \end{cases} \quad (1)$$

Where $S_{\theta_1}$ and $C_{\theta_2+\theta_3}$ denotes $\sin(\theta_1)$ and $\cos(\theta_2 + \theta_3)$, respectively. In the following, we present the nonlinear dynamic model of the quadruped robot leg using the Euler-Lagrange formulation, employing the generalized coordinates $q = [\theta_1, \theta_2, \theta_3]^T$. The detailed derivation of the dynamics model can be found in [8].

$$\mathbf{M}(q)\ddot{q} + \mathbf{V}(q, \dot{q}) + \mathbf{G}(q) = \tau + \mathbf{J}_t^T f_t \quad (2)$$

Where $\mathbf{M}(q)$ denotes the 3×3 mass matrix, $\mathbf{V}(q,\dot{q})$ is the 3×1 vector representing the centrifugal and Coriolis terms, $\mathbf{G}(q)$ is the 3×1 vector includes the gravity terms, $\tau$ is the 3×1 vector of control torques, and $\mathbf{J}_t^T f_t$ is the external torque applied to the foot tip.

## III. Controller Strategies

In this section, a number of some selected control methods for the task of trajectory tracking are provided. First, the most commonly used robust controller, the SMC technique is discussed. Then, two model-free control algorithms, including the Transpose Jacobian (TJ) and Adaptive Transpose Jacobian (ATJ) are introduced.

TABLE I. Geometric and Mass Parameters of the Robot

| Link | Lengths and Inertia Parameters | | | |
|---|---|---|---|---|
| | $m$ (kg) | $l$ (m) | $l_C$ [a] (m) | $I$ (kg.m$^2$) |
| 1 | 0.10 | 0.12 | 0.060 | 1.20×10$^{-4}$ |
| 2 | 0.30 | 0.36 | 0.180 | 3.24×10$^{-3}$ |
| 3 | 0.15 | 0.36 | 0.175 | 1.62×10$^{-3}$ |

[a.] The Length of the Center of Mass

TABLE II. Denavit-Hartenberg Parameters of the Robot

| Link, i | D-H Parameters | | | |
|---|---|---|---|---|
| | $a_{i-1}$ | $\alpha_{i-1}$ | $d_i$ | $\theta_i$ |
| 1 | 0 | 0 | 0 | $\theta_1$ |
| 2 | 0 | $-\pi/2$ | $l_1$ | $\theta_2$ |
| 3 | $l_2$ | 0 | 0 | $\theta_3$ |
| 4 | $l_3$ | 0 | 0 | 0 |

### A. Sliding Mode Control (SMC)

In order to design the SMC method, we rewrite (2) as follows:

$$\mathbf{M}(q)\ddot{q} = F + u \quad (3)$$

Where $F = \mathbf{J}_t^T f_t - \mathbf{V}(q,\dot{q}) + \mathbf{G}(q)$ and $u = \tau$. Now, using equation (3), the acceleration can be calculated as follows:

$$\ddot{q} = \mathbf{M}(q)^{-1}(F + u) \quad (4)$$

The sliding surfaces are defined as a weighted linear combination of position tracking error $\tilde{q} = q - q_d$, and velocity tracking error $\dot{\tilde{q}}$, where the subscript d denotes the desired trajectory. Therefore, the sliding surfaces are defined as below:

$$s = \dot{\tilde{q}} + \lambda \tilde{q} \quad (5)$$

Where $\lambda$ represents a 3×3 diagonal matrix with positive entries. Now, to analyze the controller's stability, we express the candidate Lyapunov function as follows:

$$\Gamma(q, \dot{q}) = \tfrac{1}{2} s^T s \quad (6)$$

Therefore, the stability of the control system requires the negative definiteness of the time derivative of the Lyapunov function. In other words, we have:

$$\tfrac{1}{2} \tfrac{d}{dt} s_i^2 = s_i \dot{s}_i \leq -\eta_i |s_i|, \quad i = 1,2,3 \quad (7)$$

Where $\eta_i$ is a positive constant parameter expressing how fast the trajectories will reach the sliding surface $s_i = 0$. The control law is obtained by differentiating the sliding surface equations and setting $\dot{s} = 0$. By substituting from (4) and adding the sign functions in order to increase the robustness of the controller, we will have:

$$u = -\widehat{\mathbf{M}}_s^{-1} \big[ \widehat{F}_s + \dot{s}_r + \mathbf{K}\,\mathrm{sgn}(s) \big] \quad (8)$$

Where $\dot{s}_r = -\ddot{q}_d + \lambda \dot{\tilde{q}}_r$, and the matrices $\widehat{\mathbf{M}}_s$ and $\widehat{F}_s$ represent the estimated values of $\mathbf{M}_s$ and $F_s$, respectively, which are defined as follows

$$\begin{aligned} \mathbf{M}_s &= \mathbf{M}^{-1} \\ F_s &= \mathbf{M}^{-1} F \end{aligned} \quad (9)$$

Also,

$$\mathbf{K}\,\mathrm{sgn}(s) = [k_1 \mathrm{sgn}(s_1), k_2 \mathrm{sgn}(s_2), k_3 \mathrm{sgn}(s_3)]^T \quad (10)$$

The reaching conditions in (7) are satisfied if the gains of the sign functions $\mathbf{K}$ are defined based on the dynamic uncertainty bound $F_{un,s}$:

$$\mathbf{K} \geq F_{un,s} + \eta, \quad |F_s - \widehat{F}_s| \leq F_{un,s} \quad (11)$$

Fig. 2 depicts the block diagram of the SMC strategy.

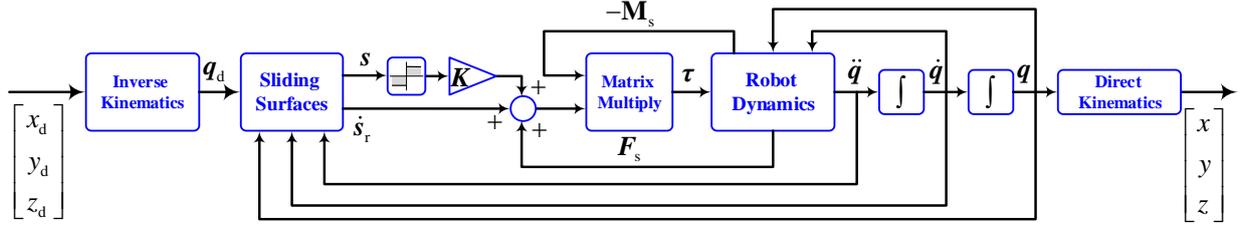

Fig. 2. Block diagram of the SMC strategy.

## B. Transpose Jacobian (TJ)

The transpose Jacobian control algorithm was introduced by Craig as follows [7]:

$$\boldsymbol{\tau} = \mathbf{J}^T(\boldsymbol{q})\{\mathbf{K}_d \dot{\boldsymbol{e}}(t) + \mathbf{K}_P \boldsymbol{e}(t)\} \quad (12)$$

Where $\mathbf{J}(\boldsymbol{q})$ represents the Jacobian matrix of the leg, which is defined as below:

$$\mathbf{J}(\boldsymbol{q}) = [J_{ij}]_{3\times3}, \quad i,j = 1,2,3 \quad (13)$$

With:

$$\begin{aligned}
J_{11} &= -l_1 C_{\theta_1} - l_2 S_{\theta_1} C_{\theta_2} - l_3 S_{\theta_1} C_{\theta_2+\theta_3} \\
J_{12} &= -l_2 C_{\theta_1} S_{\theta_2} - l_3 C_{\theta_1} S_{\theta_2+\theta_3} \\
J_{13} &= -l_3 C_{\theta_1} S_{\theta_2+\theta_3} \\
J_{21} &= -l_1 S_{\theta_1} + l_2 C_{\theta_1} C_{\theta_2} + l_3 C_{\theta_1} C_{\theta_2+\theta_3} \\
J_{22} &= -l_2 S_{\theta_1} S_{\theta_2} - l_3 S_{\theta_1} S_{\theta_2+\theta_3} \\
J_{23} &= -l_3 S_{\theta_1} S_{\theta_2+\theta_3} \\
J_{31} &= 0 \\
J_{32} &= l_2 C_{\theta_2} + l_3 C_{\theta_2+\theta_3} \\
J_{33} &= l_3 C_{\theta_2+\theta_3}
\end{aligned} \quad (14)$$

Also $\mathbf{K}_P$ and $\mathbf{K}_d$ are the proportional and derivative constant control gain matrices, respectively. The vector $\boldsymbol{e}(t)$ denotes the tracking error between the desired Cartesian position $(x_d, y_d, z_d)$ and the actual position of the foot tip $(x, y, z)$, which is defined as follows:

$$\boldsymbol{e}(t) = [x_d - x, \ y_d - y, \ z_d - z]^T \quad (15)$$

And the vector $\dot{\boldsymbol{e}}(t)$ represent the velocity tracking error between the desired Cartesian velocity $(\dot{x}_d, \dot{y}_d, \dot{z}_d)$ and the actual velocity output of the foot tip, $(\dot{x}, \dot{y}, \dot{z})$.

Fig. 3 depicts the block diagram of the TJ control strategy.

## C. Adaptive Transpose Jacobian (ATJ)

In this paper, the TJ control gains $\mathbf{K}_P$ and $\mathbf{K}_d$ in (12) have been modified in such a way to have an extended version of TJ, named as adaptive transpose jacobian (ATJ). In this way, the proportional and derivative matrices in (12) are now adjustable with an adaptive PI-algorithm, as below:

$$\boldsymbol{\tau} = \mathbf{J}^T(\boldsymbol{q})\{\mathbf{K}_d(t) \dot{\boldsymbol{e}}(t) + \mathbf{K}_P(t) \boldsymbol{e}(t)\} \quad (16)$$

Where $\mathbf{K}_P(t)$ and $\mathbf{K}_d(t)$ represent the adaptive control gains, and are adjusted as the following equations:

$$\mathbf{K}_P(t) = \mathbf{K}_{P,P}(t) + \int_0^t \dot{\mathbf{K}}_{P,I}(\tau) d\tau \quad (17)$$

$$\mathbf{K}_d(t) = \mathbf{K}_{d,P}(t) + \int_0^t \dot{\mathbf{K}}_{d,I}(\tau) d\tau \quad (18)$$

With:

$$\mathbf{K}_{P,P}(t) = \boldsymbol{e}^T \boldsymbol{\Gamma}_{P,P} \boldsymbol{e} \quad (19)$$

$$\dot{\mathbf{K}}_{P,I}(t) = \boldsymbol{e}^T \boldsymbol{\Gamma}_{P,I} \boldsymbol{e} - \delta_P \mathbf{K}_{P,I}(t) \quad (20)$$

$$\mathbf{K}_{d,P}(t) = \dot{\boldsymbol{e}}^T \boldsymbol{\Gamma}_{d,P} \dot{\boldsymbol{e}} \quad (21)$$

$$\dot{\mathbf{K}}_{d,I}(t) = \dot{\boldsymbol{e}}^T \boldsymbol{\Gamma}_{d,I} \dot{\boldsymbol{e}} - \delta_d \mathbf{K}_{d,I}(t) \quad (22)$$

In (19)-(22), the diagonal matrices $\boldsymbol{\Gamma}_{P,P}, \boldsymbol{\Gamma}_{P,I}, \boldsymbol{\Gamma}_{d,P}, \boldsymbol{\Gamma}_{d,I}$, and the small positive coefficients $\delta_P$ and $\delta_d$ are the control parameters. Rewriting (20) in a discrete form gives:

$$\{\mathbf{K}_{P,I}(t) - \mathbf{K}_{P,I}(t - \Delta t)\}/\Delta t = \boldsymbol{e}^T \boldsymbol{\Gamma}_{P,I} \boldsymbol{e} - \delta_P \mathbf{K}_{P,I}(t) \quad (23)$$

Now, solving for $\mathbf{K}_{P,I}(t)$ and rearranging the terms in a recursive relation yields

$$\mathbf{K}_{P,I}(t) = \boldsymbol{e}^T \boldsymbol{\Gamma}_{P,I} \boldsymbol{e} \left(\frac{\Delta t}{1+\delta_P \Delta t}\right) + \mathbf{K}_{P,I}(t - \Delta t)\left(\frac{1}{1+\delta_P \Delta t}\right) \quad (24)$$

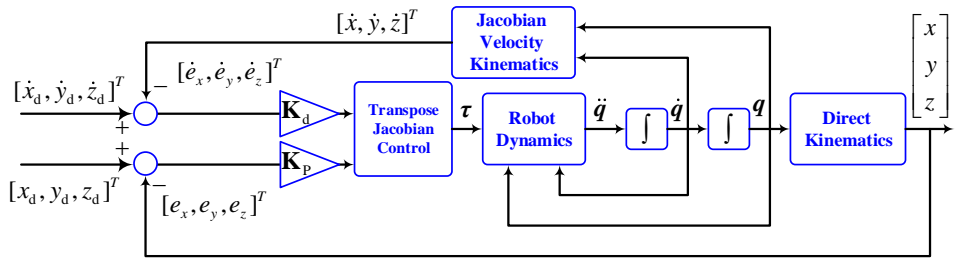

Fig. 3. Block diagram of the TJ control strategy.

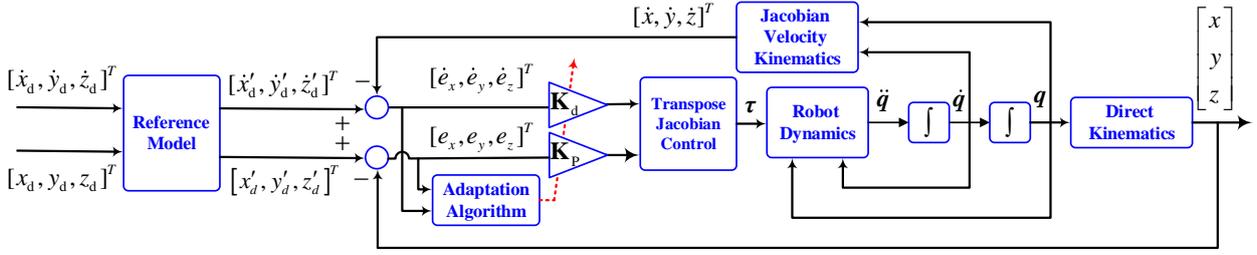

Fig. 4. Block diagram of the ATJ control strategy.

The proof of global asymptotic stability for the proposed adaptive PI-algorithm is provided in [10]. The block diagram of the ATJ control strategy is shown in Fig. 4, where an ideal reference model is proposed for error correction, which is defined as follows:

$$G(s) = \frac{\omega_n^2}{s^2 + 2\zeta\omega_n s + \omega_n^2} \quad (25)$$

The above model is the transfer function of a second-order system, which $\omega_n$ represents the natural frequency and $\zeta$ denotes the damping ratio of the system.

It should be mentioned that, in Fig. 3, the variables $x_d$, $y_d$, and $z_d$ are the desired path and $x'_d$, $y'_d$, and $z'_d$ are the corrected desired path.

## IV. PERFORMANCE EVALUATION CRITERIA

In this section, the performance of different controllers discussed earlier is evaluated. First, we introduce two criteria to evaluate and compare the performance of the controllers, then we present the simulation results for different initial conditions and uncertainty in the parameters.

### A. Assumptions

For a fair comparison of the performance of the controllers, we consider the following assumptions:

- For all control techniques, the same kinematic model of the robot is used.
- The performance of the controllers is evaluated for the same tracking path.
- No external torque is applied to the tip of the foot during the motion (*swing up phase*).
- A sigmoid function has been used for the SMC instead of the sign function, to eliminate the chattering phenomenon.

### B. Evaluation Criteria

The criteria considered to evaluate the performance of controllers include root mean square of the errors (RMSE) and the control input energy [1]:

$$\text{RMSE} = \sqrt{\left(\int_{t_i}^{t_f} e^2(\tau)d\tau\right)/(t_f - t_i)} \quad (26)$$

$$E = \int_{t_i}^{t_f} |\boldsymbol{\tau}^T(\xi)\dot{\boldsymbol{q}}(\xi)|d\xi \quad (27)$$

### C. Tuning Parameters

Due to the fact that the performance of each of the controllers is dependent on the tuning parameters, for a fair comparison of the controllers, we evaluate their performance by considering the above two criteria. The tuning parameters of each controller are shown in Table III. Also, for the ideal reference model $\omega_n = 100$ rad/s and $\zeta = 0.9$ are considered.

### D. Reference Trajectory

The trajectory needs to satisfy specific constraints to ensure that the robot walks smoothly and in a stable manner. For example, the tracking path and velocity should be continuous and differentiable. These constraints are very important for precise control of robot locomotion. Therefore, a trapezoidal curve is used in path planning [9]. By using this curve, the motion of the foot includes the process of acceleration, constant speed, and deceleration. Equation (28) expresses the trapezoidal curve. The desired path is planned based on this curve. In addition, by setting different speeds and acceleration, different curves can be produced.

$$s(t) = \begin{cases} \frac{1}{2}at^2 & 0 \leq t \leq t_a \\ vt - \frac{v^2}{2a} & t_a \leq t \leq t_f - t_a \\ vt_f - \frac{v^2}{a} - \frac{1}{2}a(t - t_f)^2 & t_f - t_a \leq t \leq t_f \end{cases} \quad (28)$$

Fig. 5 shows the trapezoidal curve for the time interval $t_f = 3$ s.

Now, using the obtained curve, the tracking path is designed as below:

$$\begin{aligned} x_d(t) &= \frac{S}{2}\cos\left(\pi\left[1 - \frac{s(t)}{s(t_f)}\right]\right) \\ y_d(t) &= \text{constant} \qquad t \in [0, t_f] \quad (29) \\ z_d(t) &= H\sin\left(\pi\left[1 - \frac{s(t)}{s(t_f)}\right]\right) \end{aligned}$$

Where $S$ is the length of the step in the horizontal direction, respectively, and $H$ is the height of the step, with no lateral motion.

## V. SIMULATION RESULTS

In this section, the simulation results are presented for the initial conditions $\boldsymbol{x}_0 = -[0.63, -0.124, 0.112]^T$ m, and all of the designed control algorithms have proven to be effective as shown in Fig. 6. The SMC and ATJ control algorithms perform well when tracking the desired path, while the TJ approach presents an overshoot.

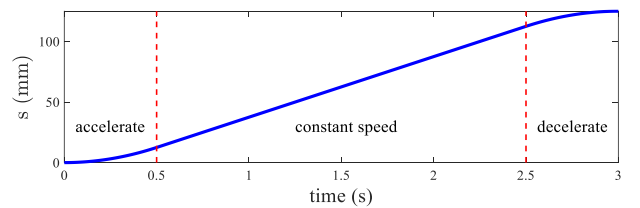

Fig. 5. Trapezoidal curve with $t_a = 0.5$ s, $a = 0.1$ m/s$^2$.

TABLE III. PARAMETERS OF THE CONTROL STRATEGIES

| Control Strategy | Tuning Parameters |
|---|---|
| SMC | $\lambda = \text{diag}(10,10,10)$<br>$\eta_1 = \eta_2 = \eta_3 = 10$ |
| TJ | $\mathbf{K}_P = \text{diag}(700,700,700)$<br>$\mathbf{K}_d = \text{diag}(7,7,7)$ |
| Adaptive TJ | $\mathbf{\Gamma}_{P,P} = \mathbf{\Gamma}_{P,I} = \text{diag}(20000,20000,40000)$<br>$\mathbf{\Gamma}_{d,P} = \mathbf{\Gamma}_{d,I} = \text{diag}(300,3000,200)$<br>$\delta_P = \delta_d = 0.04$ |

The RMSE and control input energy of the trajectory tracking task for all of the designed control algorithms are shown in Fig. 7. The SMC's tracking error and control input energy are less than other control algorithms due to its model-based algorithm. On the other hand, although the ATJ has a RMSE close to TJ, it has a remarkable performance in terms of control input energy.

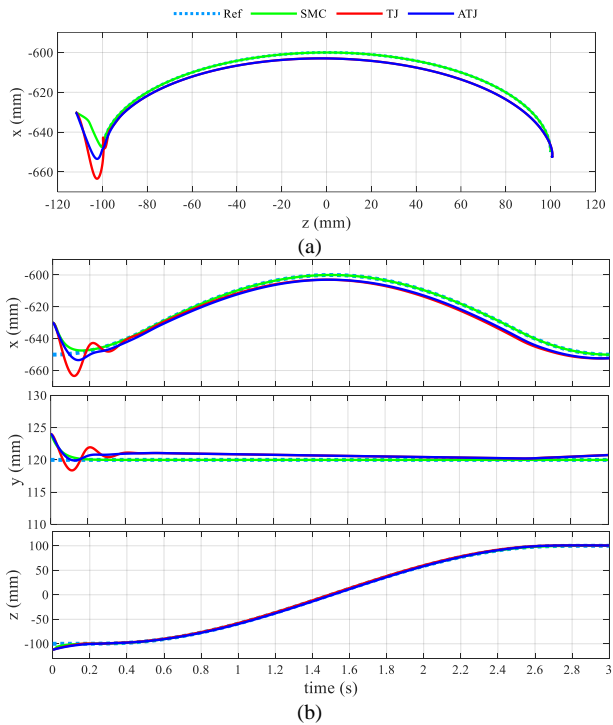

Fig. 6. Trajectory tracking results for different controllers in the time interval $t \in [0,3]$ s, for initial conditon $\boldsymbol{x}_0 = -[0.63, -0.124, 0.112]^T$ m, and desired intial condition $-[0.65, -0.12, 0.1]^T$ m. (a) Trajectory in x-z plane, (b) Time history of x, y, and z.

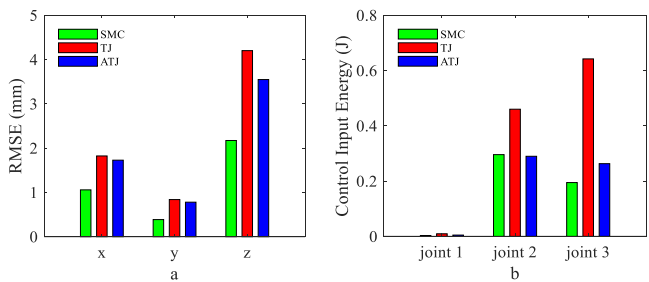

Fig. 7. (a) RMS of the errors, (b) Control input energy. initial condition: $\boldsymbol{x}_0 = -[0.63, -0.124, 0.112]^T$ m.

## A. Initial Deviation

Fig. 8 depicts the trend of the changes in RMSE for the initial deviations considered in Table IV. The results show that TJ and Adaptive TJ control algorithms do not show much sensitivity to changes in the initial conditions. On the other hand, although the SMC shows a lower RMSE compared to TJ and ATJ algorithms, it demonstrates significant changes in RMSE for the initial deviations. This highlights the greater reliance of SMC on the initial condition in contrast to TJ and ATJ control algorithms.

## B. Parameter Mismatch

The effect of uncertainty in parameters on the performance of controllers is shown in Fig. 9. Although SMC is considered a robust controller due to its model-based control algorithm, this controller can withstand up to 20% uncertainty in parameters, whereas other controllers still perform well.

## VI. CONLUSION

This study evaluated and compared the performance of three control algorithms for tracking a semi-elliptical path by a quadruped robot leg. These control strategies were chosen based on their popularity in the field of path-tracking control and their applicability to legged robots. These control techniques include sliding mode control (SMC), transpose Jacobian (TJ), and adaptive transpose Jacobian (ATJ). Two criteria of control input energy and root mean square of errors were considered to evaluate the performance of the controllers. The obtained results revealed that:

- The adaptation algorithm improves the performance of the TJ control algorithm and reduces the overshoot during path tracking.

- SMC has lower RMSE and control energy compared to other control algorithms. And although ATJ has a similar RMSE to TJ, it has low control energy.

- TJ and ATJ control algorithms are less sensitive to changes in initial conditions, while SMC shows significant changes in RMSE for initial deviations, indicating its greater reliance on initial conditions.

- SMC showed good performance up to 20% due to its model-based algorithm, whereas the TJ and ATJ control algorithms performed better even in higher uncertainties.

As a future work, due to the fact that the ATJ algorithm is a model-free control algorithm, it can be applied to quadruped robots and humanoid robots that have complex dynamic models. This will be very useful for real-time control of these types of robots. Furthermore, the ATJ algorithm can be extended to estimate the parameters for better performance in unknown environments.

TABLE IV. INITIAL DEVIATION FROM REFERENCE PATH

| Initial Conditions | Initial Deviations | | |
|---|---|---|---|
| | $\Delta x$ (mm) | $\Delta y$ (mm) | $\Delta z$ (mm) |
| $\Delta_0$ | 0 | 0 | 0 |
| $\Delta_1$ | 5 | 1 | -3 |
| $\Delta_2$ | 10 | 2 | -6 |
| $\Delta_3$ | 15 | 3 | -9 |
| $\Delta_4$ | 20 | 4 | -12 |
| $\Delta_5$ | 25 | 5 | -15 |

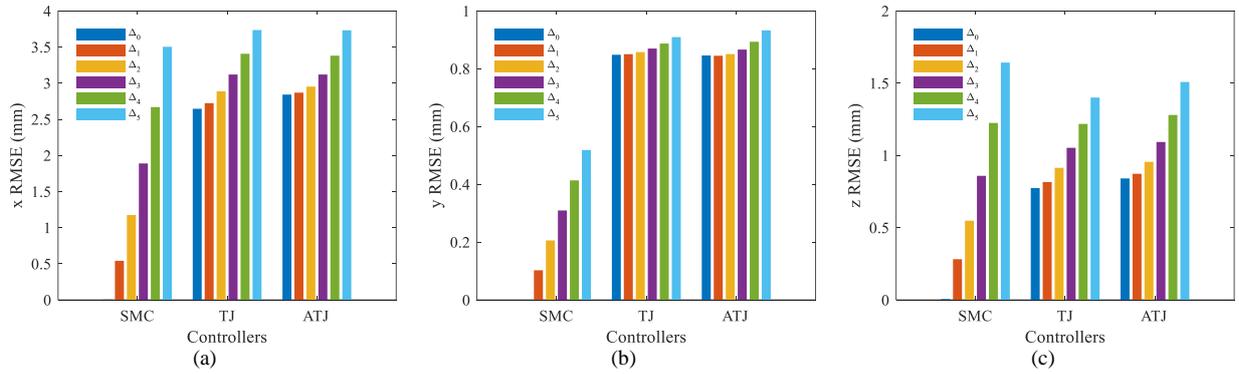

Fig. 8. RMSE of trajectory tracking for different initial conditions presented in Table IV. (a) x-direction, (b) y-direction, (c) z-direction.

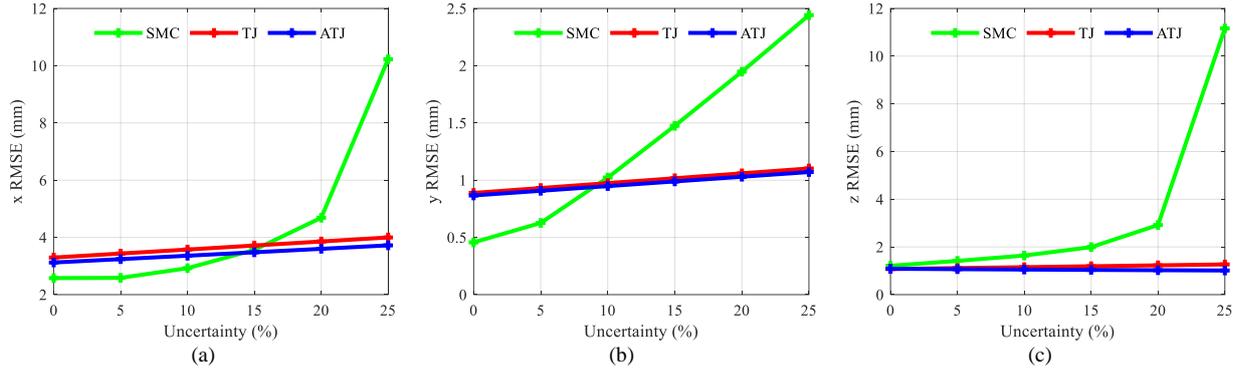

Fig. 9. RMSE of trajectory tracking for different percentage of uncertainties in mass parameter from zero to 25%. (a) x-direction, (b) y-direction, (c) z-direction.